\documentclass[runningheads]{llncs}
\usepackage{color,soul}
\usepackage{xcolor}
\usepackage{graphicx}
\usepackage{textcomp}
\usepackage{caption}
\usepackage{subcaption}
\usepackage{hyperref}
\usepackage{url}
\usepackage{array,multirow,makecell}
\usepackage{breakurl}
\usepackage{mathtools}
\usepackage{tikz}
\usepackage[square,sort,comma,numbers]{natbib}

\usepackage{lipsum}
\usepackage[title]{appendix}

\usepackage{amsmath,amssymb}

\usepackage{enumitem}
\usepackage{booktabs}
\usepackage{soul}
\usepackage{multirow}

\definecolor{light-red}{HTML}{FF7E71}
\definecolor{light-blue}{HTML}{64A1FF}
\definecolor{light-yellow}{HTML}{FFCC71}

\newif\ifconsiderlater
\considerlatertrue

\ifconsiderlater
	\newcommand{\todo}[1]{\textcolor{blue}{\textbf{XXX [#1] XXX}}}
\else
\newcommand{\todo}[1]{}
\fi

\begin{document}
\title{Diffusion-based Data Augmentation for Skin Disease Classification: Impact Across Original Medical Datasets to Fully Synthetic Images}
%

\author{Mohamed Akrout\inst{1}$^*$ \and
Bálint Gyepesi\inst{1}$^*$ \and Péter Holló\inst{2} \and Adrienn Poór\inst{2} \and\\
Blága Kincső\inst{2}\and Stephen Solis\inst{1} \and Katrina Cirone\inst{1} \and Jeremy Kawahara\inst{1} \and \\ Dekker Slade \inst{1} \and Latif Abid\inst{1} \and Máté Kovács\inst{1} \and István Fazekas\inst{1}}

\titlerunning{Diffusion-based Image Augmentation for Skin Disease Classification}
\authorrunning{M. Akrout, B. Gyepesi et al.}
%
\institute{AIP Labs, Budapest, Hungary \and Semmelweis University, Faculty of Medicine, Department of Dermatology, Venereology and Dermatooncology, Budapest, Hungary
}
%
\maketitle              
\def\thefootnote{*}\footnotetext{equal contribution}

\vspace{-0.2cm}
\begin{abstract}
Despite continued advancement in recent years, deep neural networks still rely on large amounts of training data to avoid overfitting. However, labeled training data for real-world applications such as healthcare is limited and difficult to access given longstanding privacy, and strict data sharing policies. By manipulating image datasets in the pixel or feature space, existing data augmentation techniques represent one of the effective ways to improve the quantity and diversity of training data. Here, we look to advance augmentation techniques by building upon the emerging success of text-to-image diffusion probabilistic models in augmenting the training samples of our macroscopic skin disease dataset. We do so by enabling fine-grained control of the image generation process via input text prompts. We demonstrate that this generative data augmentation approach successfully maintains a similar classification accuracy of the visual classifier even when trained on a fully synthetic skin disease dataset. Similar to recent applications of generative models, our study suggests that diffusion models are indeed effective in generating high-quality skin images that do not sacrifice the classifier performance, and can improve the augmentation of training datasets after curation.


\end{abstract}
\vspace{-0.7cm}
\begin{figure}[!h]
  \centering
  \includegraphics[scale=0.4]{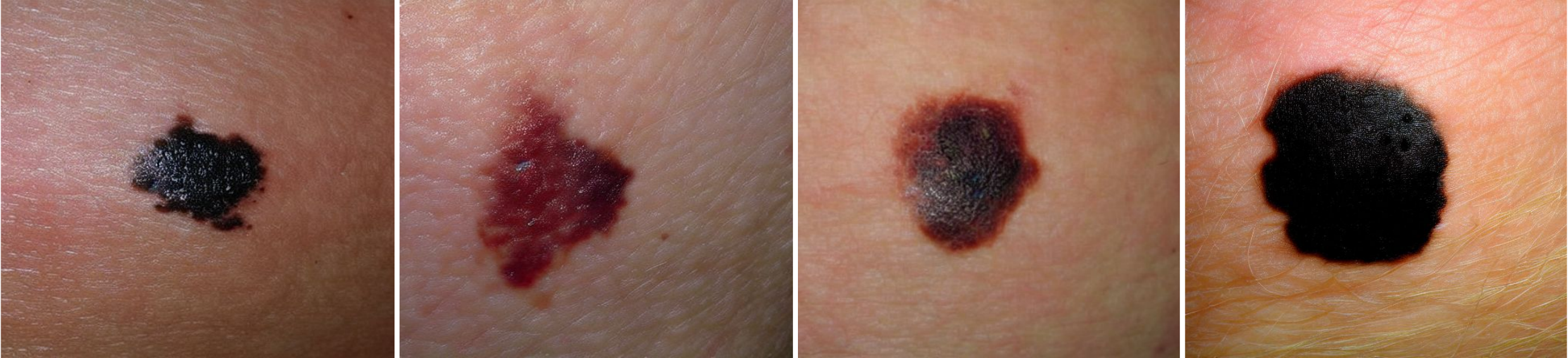}
  \caption{Synthetic melanoma images generated by the stable diffusion model after fine-tuning it with melanoma images using the input text prompt ``melanoma''.}
  \label{fig:synthetic-melanoma}
\end{figure}

\section{Introduction}

The last months have witnessed the emergence of diffusion probabilistic models (DPM) \cite{ho2020denoising} as a powerful generator of high-fidelity synthetic datasets, leading to record-breaking performances in various applications such as image synthesis \cite{rombach2022high}, natural language processing \cite{austin2021structured}, and computational chemistry \cite{anand2022protein}, to name a few. When compared to other types of generative models, such as generative adversarial networks (GANs) and variational autoencoders, DPMs are easier to train and offer state-of-the-art image generation quality \cite{croitoru2022diffusion}.

Given that synthetic images play a crucial role in privacy-preserving generation and small dataset augmentation, DPMs attracted significant attention in the medical imaging field. Table~\ref{fig:params-MC-model} provides an overview of the prior studies of DPMs, including their medical applications  and dataset domains. At first glance, the reader can identify that the study in \cite{sagers2022improving} is the closest one to this work where synthetic images were generated from seed images in the Fitzpatrick 17k dataset using the
OpenAI's DALL·E 2 model \cite{ramesh2022hierarchical}.\vspace{-0.55cm} 

\begin{table}
\centering
\caption{Summary of existing applications of diffusion models in medical imaging.}
\vspace{0.1cm}
\begin{tabular}[t]{c @{\hskip 14\tabcolsep} c @{\hskip 14\tabcolsep} c}
\toprule
\multirow{2}{*}{\textbf{Medical applications}} & \multirow{2}{*}{\textbf{Dataset domain}} & \multirow{2}{*}{\textbf{Papers}}\\\\
\midrule
Image generation & lungs X-Ray, CT, MRI & {\cite{packhauser2022generation,ali2022spot,chambon2022adapting,pinaya2022brain}}  \\ \midrule
Image segmentation & MRI, CT, ultrasound & { \cite{wu2022medsegdiff,guo2022accelerating,la2022anatomically}}  \\ \midrule
Image inpainting & MRI& {\cite{rouzrokh2022multitask}} \\ \midrule
Image denoising & MRI, CT, retinal OCT & {\cite{xia2022low,chung2022mr,hu2022unsupervised}} \\ \midrule
Lesion detection & MRI & {\cite{sanchez2022healthy,wyatt2022anoddpm,wolleb2022diffusion}}  \\ \midrule
Image translation & MRI, CT & {\cite{ozbey2022unsupervised,la2022anatomically}}\\ 
\midrule
Seed-image based augmentation & Dermatology & \cite{sagers2022improving}\\
\midrule
\multirow{2}{*}{\textbf{Skin disease classification}} & \multirow{3}{*}{\textbf{Dermatology}} & \multirow{3}{*}{\textbf{This work}}\\
\multirow{2}{*}{\textbf{using large synthetic datasets}} &  &   \\
 &  &   \\
\bottomrule
\end{tabular}
\label{fig:params-MC-model}
\vspace{-0.2cm}
\end{table}

\noindent Inspired by the recent early success of DPMs, we propose to use diffusion models for image augmentation as part of supervised machine learning pipelines. More specifically, we study how diffusion models can $i)$ increase the classification metrics for skin diseases, and $ii)$ augment skin condition datasets by effectively manipulating the generated images' features conditioned on the input text prompts. This paper makes the following contributions:
\begin{itemize}
    \item We study the potential of DPMs for skin disease classifications by fine-tuning them on six different disease conditions: basal cell carcinoma, melanoma, actinic keratosis, atypical melanocytic nevus, lentigo, seborrheic keratosis. We do so by learning the embeddings of each disease using text inversion.
    \item We demonstrate that the classification accuracies of skin disease classifiers trained on generated synthetic images is similar to training on real images, where the performance is maintained when using half the number of real images, and only slightly deteriorates when using a fully synthetic dataset. This result suggests that the recent success of generative models can help minimize the barriers of sharing labeled medical datasets, with minimal performance deterioration.
    \item We illustrate how DPMs are powerful tools to add visual aspects of skin images guided by domain experts in complementing training datasets.
\end{itemize}


\section{Diffusion-based data augmentation}
In this section, we begin by describing the methods used for training the embeddings of the aforementioned six skin diseases on our macroscopic skin images. Then, we present the datasets associated with the two DPM training scenarios: a hybrid dataset compromising 50\% synthetic and 50\% real images, and a 100\% fully synthetic dataset generated by the trained embeddings.

\subsection{Stable diffusion}\label{subsec:stable-diffusion}

\begin{figure}[!h]
  \centering
  \includegraphics[scale=0.37]{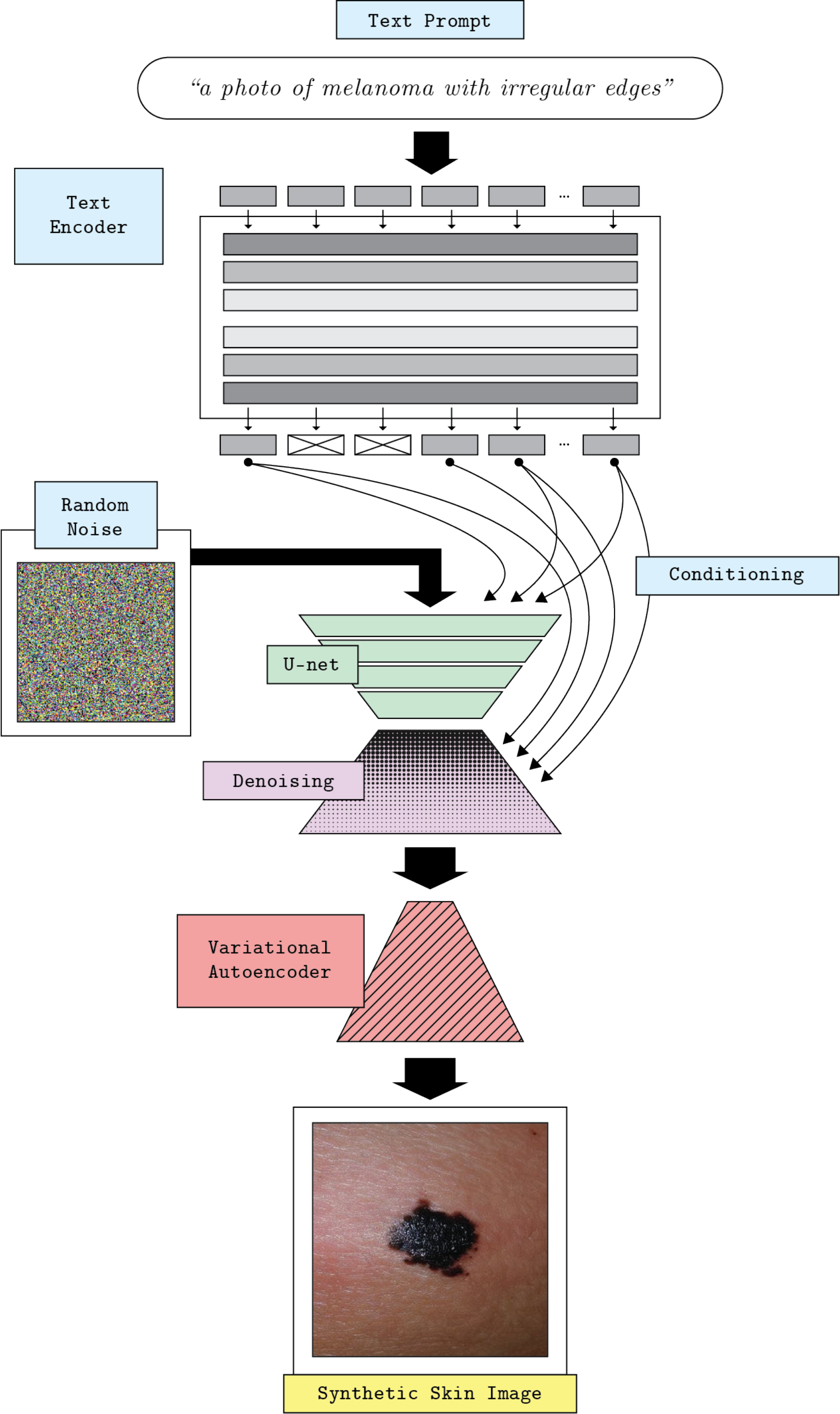}
  \caption{The diffusion model pipeline for synthetic skin image generation.}
  \label{fig:stable-diffusion-pipeline}
  \vspace{-0.5cm}
\end{figure}

The stable diffusion model proposed in \cite{rombach2022high} is not a monolithic model, but rather a pipeline of three components, as depicted in Fig. \ref{fig:stable-diffusion-pipeline}:
\begin{itemize}
    \item[$1)$]  \textit{Text encoding}, based on the CLIP model \cite{radford2021learning}, which transforms each token of the input text prompt into an embedding vector. 
    \item[$2)$] \textit{Latent space U-Net generator}, which takes all the token embeddings and a random noise array (a.k.a., latent array) and sequentially generates multiple arrays that better resemble the input text and the visual images on which the U-Net has been trained.
    \item[$3)$] \textit{Image decoder}, based on a variational autoencoder (VAE) to transform the obtained latent array into the pixel space.
\end{itemize}
\noindent In this pipeline, the embedding vectors of the text encoding control both the generation of the U-Net latent space representations as well as the VAE decoding.

\vspace{-0.2cm}
\subsection{Training dataset for synthetic image generation}
The limited number of available labeled images is one of the leading limitations faced by medical classification applications. Our internal macroscopic image dataset consists of thousands of skin condition images curated and classified by dermatologists to cover more than 700 different diseases. Here, we choose six widely spread classes across three distinct categories:
\begin{itemize}[leftmargin=*]
    \item \textit{Malignant classes}: basal cell carcinoma and melanoma;
    \item \textit{Pre-malignant classes}: actinic keratosis and atypical melanocytic nevus;
    \item \textit{Benign classes}: lentigo and seborrheic keratosis.
\end{itemize}


\noindent Table \ref{tab:internal-dataset} provides an overview of the number of images used for each disease in training the text embedding with the stable diffusion model.

\begin{table*}[!bh]
\vspace{-0.4cm}
\caption{The number of real training images for the considered skin diseases.}
\vspace{0.1cm}
\centering
\begin{tabular}{ccc}
\toprule
    \textbf{Category} & \textbf{Skin disease} & \textbf{Data source}\\

    \midrule
        \multirow{2}{*}{Benign} & {Seborrheic keratosis} & 2134 \\
    & {Lentigo} & 680  \\
    \midrule
    \multirow{2}{*}{Pre-malignant} & {Actinic keratosis} & 3298  \\
    & {Atypical melanocytic nevus} & 623 \\
    \midrule
    \multirow{2}{*}{Malignant} & Basal cell carcinoma & 7081  \\
    & {Melanoma} & 3381  \\
    \bottomrule
\end{tabular}
\label{tab:internal-dataset}
\end{table*}

\noindent In order to train the text embeddings associated to each skin disease, we use the stable diffusion architecture \cite{rombach2022stablediffusion} based on latent diffusion models \cite{rombach2022high}. Using a model of the latter pretrained on multiple LAION datasets \cite{Laion}, we fine-tune each embedding on our real-world image skin condition dataset for two million steps using the default hyperparameters proposed in \cite{stein2022invokeai}. We use PyTorch for both training and inference. Each embedding is trained on three NVIDIA GeForce RTX 3090 GPUs.

\subsection{Curation of generated images}\vspace{-0.1cm}
While most of the generated skin disease images are of high quality, it is not unusual to obtain generated images of medium or low quality. To isolate high-quality images from lower qualities, Fig.\ref{fig:generation-pipeline} depicts the full pipeline for augmenting our skin disease dataset composed of the following four steps:\vspace{-0.35cm}

\begin{figure}[!h]
  \centering
  \includegraphics[scale=0.35]{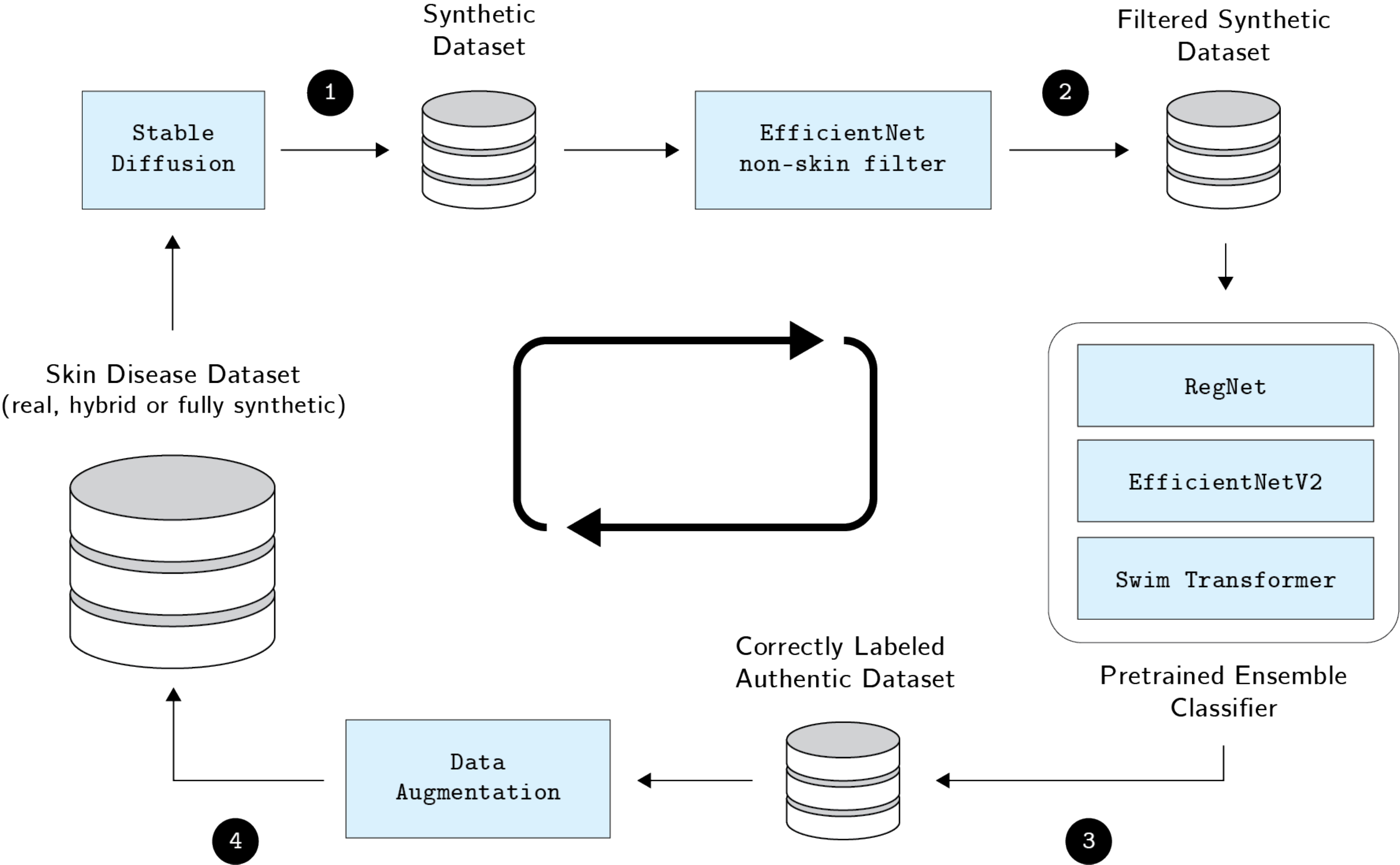}
  \caption{Summary of the four steps of the generation pipeline for skin disease data augmentation.}
  \label{fig:generation-pipeline}
  \vspace{-0.6cm}
\end{figure}

\begin{itemize}[leftmargin=*]
    \item[$1)$] \textit{Synthetic data generation}: Using the stable diffusion model described in Section \ref{subsec:stable-diffusion}, we generate 30.000 images for each one of the considered six skin diseases to get a synthetic dataset.
    \item[$2)$] \textit{Non-skin image filtering}: We run the obtained synthetic dataset in $1)$ through a pretrained binary EfficientNet classifier \cite{tan2020efficientnet} to filter out any non-skin images. The binary classifier has been trained on the skin images of the macroscopic dataset presented in Table~\ref{tab:internal-dataset} and non-skin images from ImageNet. The accepted images as skin images by the binary classifier represent more than $99\%$ of the generated images and constitute the filtered synthetic dataset.
    \item[$3)$] \textit{Skin disease image filtering}: We use the filtered synthetic dataset to predict the skin disease label using a pretrained ensemble model composed of two CNN models (EfficientNetV2 \cite{tan2021efficientnetv2}, RegNet \cite{dollar2021regnet}) and a visual transformer (Swin-Transformer \cite{liu2021swin}). This ensemble model has been pretrained on the macroscopic dataset presented in Table~\ref{tab:internal-dataset}.
    \item[$4)$] \textit{Data augmentation}: We use the correctly labeled images by the pretrained ensemble classifier as the data source for augmenting our initial dataset.\vspace{-0.2cm}
\end{itemize}

\section{Experiments and Results}
\subsection{Dataset scenarios for synthetic image generation}\label{subsec:datasets-for-synthetic-images}
Based on the filtered images whose labels were correctly predicted by the pretrained ensemble classifier, we build a fully synthetic dataset consisting of 500 images per skin disease. For the real images, we randomly sample 500 images per class from our macroscopic skin image dataset. To examine the impact of the synthetic dataset on classification metrics, we consider the following datasets:
\begin{itemize}[leftmargin=*]
    \item a \textit{small real dataset} (real-small) containing 250 real images only,
    \item a \textit{real dataset} containing 500 real images only,
    \item a \textit{hybrid dataset} consisting of 250 real images and 250 synthetic images,
    \item a \textit{synthetic dataset} containing 500 synthetic images only.    
\end{itemize}

\noindent Note that the four datasets are balanced across skin diseases with varying proportions of real and synthetic images. This allows us to assess the efficiency of substituting real data with synthetic ones.



\subsection{Medical synthetic data samples using text prompt inputs}
\begin{figure}[!h]
  \centering
  \includegraphics[scale=0.34]{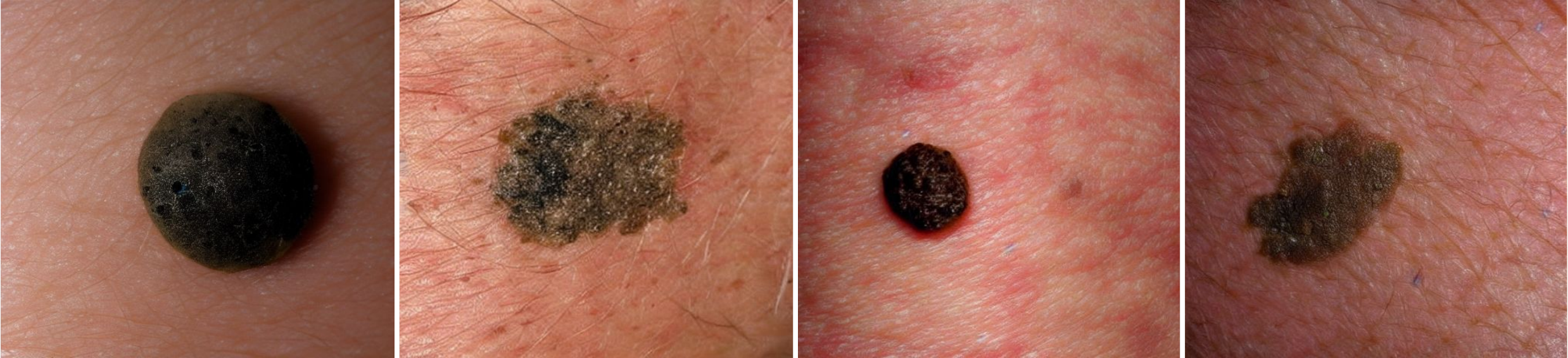}
  \caption{Synthetic seborrheic keratosis images generated by the stable diffusion model after fine-tuning it with seborrheic keratosis images using the input text prompt ``seborrheic keratosis''.}
  \label{fig:synthetic-seborrheic-keratosis}
\end{figure}

\begin{figure}[!h]
  \centering
  \includegraphics[scale=0.34]{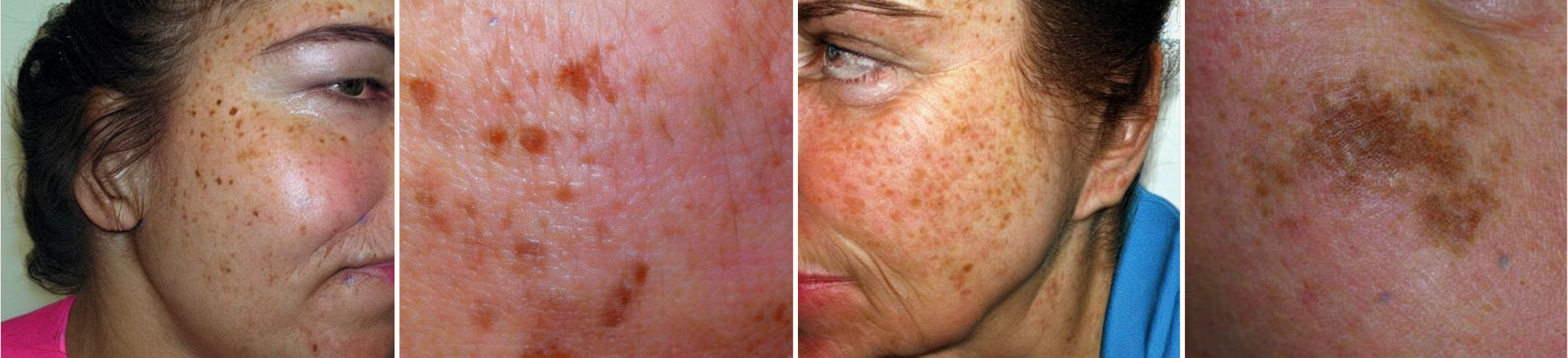}
  \caption{Synthetic lentigo images generated by the stable diffusion model after fine-tuning it with lentigo images using the input text prompt ``lentigo''.}
  \label{fig:synthetic-lentigo}
\end{figure}

\begin{figure}[!h]
  \centering
  \includegraphics[scale=0.34]{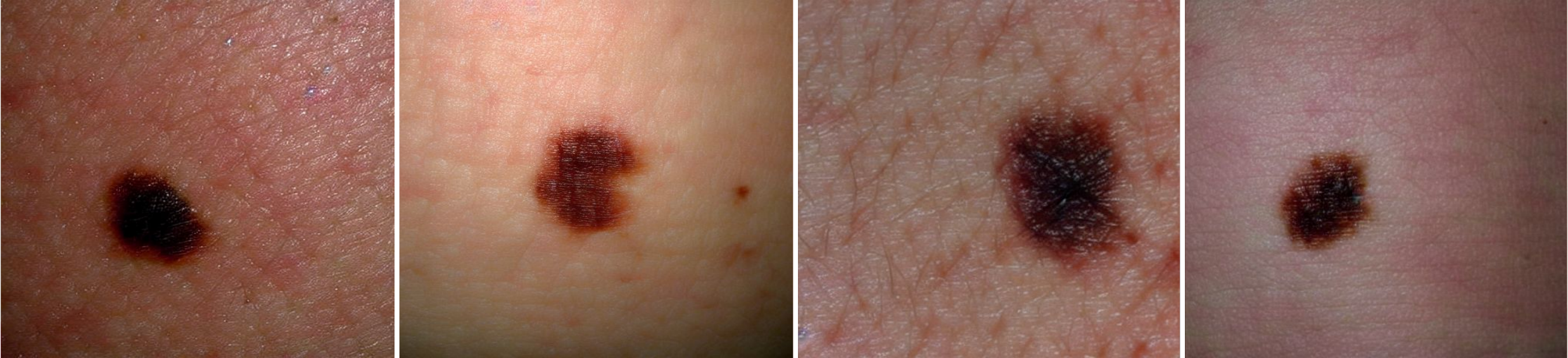}
  \caption{Synthetic synthetic atypical melanocytic nevus images generated by the stable diffusion model after fine-tuning it with atypical melanocytic nevus images using the input text prompt ``atypical melanocytic nevus''.}
  \label{fig:synthetic-melanocytic-nevus}
\end{figure}

\begin{figure}[!h]
  \centering
  \includegraphics[scale=0.34]{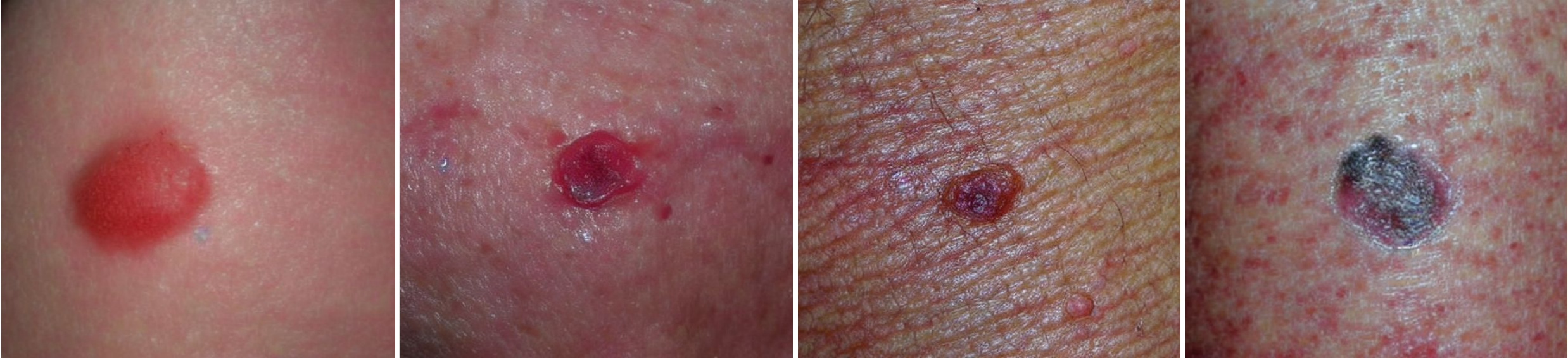}
  \caption{Synthetic basal cell carcinoma images generated by the stable diffusion model after fine-tuning it with basal cell carcinoma images using the input text prompt ``basal cell carcinoma''.}
  \label{fig:synthetic-basal-cell-carcinoma}
\end{figure}

\begin{figure}[!h]
  \centering
  \includegraphics[scale=0.34]{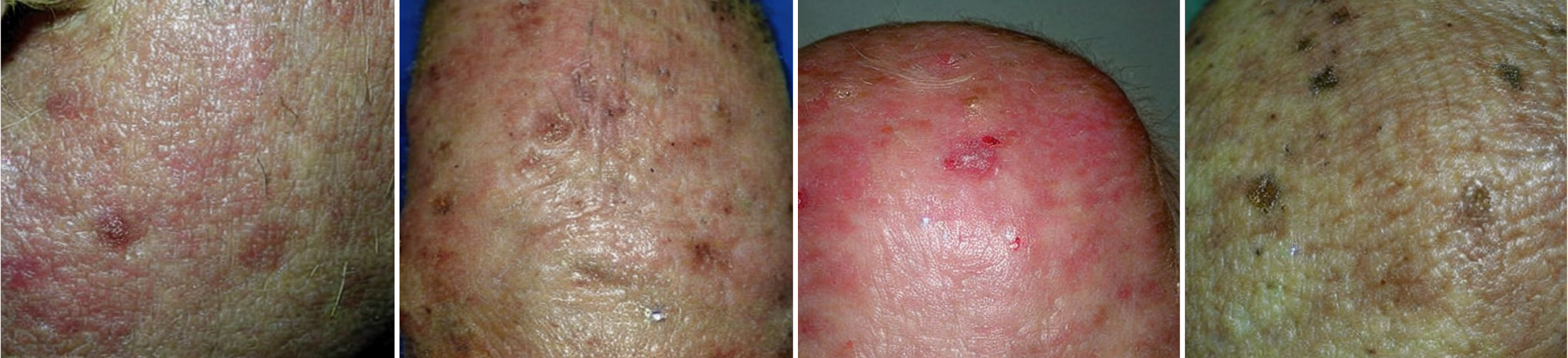}
  \caption{Synthetic actinic keratosis images generated by the stable diffusion model after fine-tuning it with actinic keratosis images using the input text prompt ``actinic keratosis''.}
  \label{fig:synthetic-actinic-keratosis}
  \vspace{-0.2cm}
\end{figure}

Here, we demonstrate the quality of the synthetic skin disease images stemming from the generation pipeline in Fig. \ref{fig:generation-pipeline} by providing four synthetic images for each disease. Similar to the synthetic melanoma images in Fig. \ref{fig:synthetic-melanoma}, we present synthetic images of seborrheic keratosis, lentigo, atypical melanocytic nevus, basal cell carcinoma and actinic keratosis in Figs. \ref{fig:synthetic-seborrheic-keratosis}, \ref{fig:synthetic-lentigo}, \ref{fig:synthetic-melanocytic-nevus}, \ref{fig:synthetic-basal-cell-carcinoma}, and \ref{fig:synthetic-actinic-keratosis}, respectively.

\vspace{-0.5cm}

\noindent While the impressive generative capabilities of AI models have already been established for normal and glaucomatous eyes in \cite{kumar2022evaluation}, our generated macroscopic images for different skin diseases similarly establishes the effectiveness for dermatology using larger synthetic datasets. This is to be opposed to seed-image based augmentation in \cite{sagers2022improving} where synthetic datasets where not used to fine-tune the generative model.

\subsection{Classification of Skin Conditions}\label{sec:classification}
In this section, we first describe the training and inference procedures of the skin disease ensemble classifier on the four datasets described in Section  \ref{subsec:datasets-for-synthetic-images}.
\subsubsection{The Training Step}
We start by training three networks of the ensemble classifier (i.e., Swin-Transformer \cite{liu2021swin}, EfficientNetV2 \cite{tan2021efficientnetv2}, and RegNetZ \cite{dollar2021regnet}) on each one of the datasets (i.e., real, hybrid, and synthetic). We do so using the PyTorch Image Models library \cite{wightman2022timm}. We make use of the default training hyperparameters and set the number of training epochs and batch size to 100 and 8, respectively. We also use early stopping\footnote{Here, early stopping occurs as soon as the validation accuracy does not improve over 10 consecutive epochs.} by monitoring the validation loss, and opt for the stochastic gradient descent (SGD) optimizer. We also use a data split of 80\% and 20\% for training and validation dataset sizes, respectively.

For every dataset, we calculate the mean and standard deviation for each one of the RBG image channels. They are accustomed to preprocessing the input images to normalize the images fed to all the networks. It is worth noting that the early stopping criterion occurs when we train the models on the fully synthetic dataset only. This is as opposed to training on real or hybrid datasets, where early stopping does not occur because the validation accuracy stagnates with very little increase, and peaks at 89\% only. This observation suggests that the fully synthetic dataset generated with stable diffusion exhibits non-perceptible differentiating features that is allowing for faster training and convergence. 

\subsubsection{The Inference Step}
We evaluate the trained ensemble model by running inference on our test dataset consisting of 3582 real images. Table \ref{tab:test-dataset} shows their distribution across the skin disease categories and classes. 

\begin{table*}[htb!]
\caption{The number of test images for the six considered skin diseases}
\vspace{0.3cm}
\centering
\begin{tabular}{ccc}
\toprule
    \textbf{Category} & \textbf{Skin disease} & \textbf{Number of images}\\
    \midrule
        \multirow{2}{*}{Benign} & {Seborrheic keratosis} & 1597 \\
    & {Lentigo} & 293  \\
    \midrule
    \multirow{2}{*}{Pre-malignant} & {Actinic keratosis} & 282  \\
    & {Atypical melanocytic nevus} & 885 \\
    \midrule
    \multirow{2}{*}{Malignant} & Basal cell carcinoma & 345  \\
    & {Melanoma} & 180  \\
    \bottomrule
\end{tabular}
\label{tab:test-dataset}
\end{table*}

\noindent We do not carry out any preprocessing to the test images other than the same normalization applied to the training images.
 
\subsection{Classification results}\label{sec:results}

We now evaluate three ensemble classifiers where each classifier is separately trained on one of the real-small, real, hybrid and synthetic datasets, as described in Section \ref{subsec:datasets-for-synthetic-images}. We run inference on our test dataset and report in Table \ref{tab:classifier-results} the associated top-k classification accuracy. The latter computes the number of times where the correct skin disease is among the top-k predicted diseases (ranked from highest to lowest predicted scores).

\begin{table}[hbt!]
\caption{Top-1 to top-5 skin disease classification accuracy on real-small, real, hybrid and fully synthetic datasets.}
\vspace{0.3cm}
\centering
  \begin{tabular}{cccccccc}
    \toprule
    \multirow{2}{*}{\textbf{Dataset}} \hspace{0.1cm} &
      \multicolumn{2}{c}{\textbf{\# of images}} &
      \multicolumn{5}{c}{\textbf{Accuracy}} \\
      \cmidrule[1pt](r{3pt}){2-3} \cmidrule[1pt](l{0.1pt}){4-8} & \textit{~Real} \hspace{0.3cm} & \textit{~Synthetic} \hspace{0.3cm} & \textit{~\,Top-1} \hspace{0.3cm} & \textit{~\,Top-2} \hspace{0.3cm} & \textit{~\,Top-3} \hspace{0.3cm} & \textit{\,~Top-4} \hspace{0.3cm} & \textit{Top-5} \hspace{0.3cm}\\
    \midrule
        Real-small & 250 & 0 & 53.41\% & 73.51\% & 83.22\% & 89.75 \%& 95.45\%\\
    \midrule
        Real & 500 & 0 & 54.05\% & \textbf{73.95}\% & 84.84\% & 91.49 \%& \textbf{96.96}\%\\
    \midrule
        Hybrid & 250 & 250 & \textbf{54.13\%} & 73.23\% & \textbf{85.01\%} & \textbf{92.16\%} & 96.65\%\\
    \midrule
        Synthetic & 0 & 500 & 47.29\% & 70.71\% & 84.09\% & \textbf{92.16}\% & 96.85\%\\
    \bottomrule
  \end{tabular}
\label{tab:classifier-results}
\end{table}

\noindent From Table \ref{tab:classifier-results}, it can be seen that the top-k accuracies of the four classifiers are very comparable. More importantly, we observe how the use of synthetic images improves the overall accuracy  of skin classifiers. Indeed, their performances on the real and hybrid datasets have been improved. As ascertained by our clinical partners at Semmelweis University, this result confirms that beyond their impressive visual quality across thousands of images, diffusion models also provide significant benefit as synthetic images for real-world medical applications.


\section{Conclusion}\label{sec:conclusions}
In this paper, we demonstrate the impressive generative capabilities of probabilistic diffusion models in generating macroscopic skin disease images. We show how it is possible to condition the probabilistic diffusion-based generation on text prompt inputs in obtaining fine-grained synthetic images. Furthermore, we propose a closed loop data augmentation pipeline to automatically curate the generated images while complementing real-world skin disease datasets. Finally, our classification task of six skin diseases highlights how synthetic images are reliable data sources given that they have been demonstrated beneficial for skin disease classification. This result underlines the importance of the recent generative modelling success for medical applications as an effective means of data sharing without infringing confidentiality issues. Several exciting avenues for further investigation remain open such as conditioning the image generation in relation to skin tone, with skin tone diversification in datasets being another leading limitation, or the use of input images in addition to the text prompt.




\let\oldbibliography\bibliography
\renewcommand{\bibliography}[1]{{%
  \let\chapter\section
  \oldbibliography{#1}}}

{
\bibliographystyle{splncs04}
\bibliography{micaai2023.bib}
}
\end{document}